\documentclass{article} 
\usepackage{iclr2026,times}

\newcommand{\method}{FlashDrive\xspace}

\newcommand{\trpt}[1]{\textcolor{NavyBlue}{\scriptsize #1}}

\usepackage{color}
\usepackage[dvipsnames]{xcolor}
\usepackage{epsfig}
\usepackage{graphicx}

\usepackage{adjustbox}
\usepackage{array}
\usepackage{booktabs}
\usepackage{colortbl}
\usepackage{float,wrapfig}
\usepackage{multirow}
\usepackage{subcaption} 
\usepackage[toc,page]{appendix}
\usepackage{stfloats}
\usepackage{paralist}
\usepackage{tabularx}
\usepackage{xspace} 
\usepackage{amsmath,amsfonts,amsthm,amssymb}
\usepackage{bm}
\usepackage{nicefrac}
\usepackage{microtype}
\usepackage{inconsolata}
\usepackage{pifont}
\usepackage{bbm}
\usepackage{booktabs}
\usepackage{cuted}
\usepackage{enumitem}
\usepackage{wrapfig}

\usepackage[pagebackref,breaklinks,colorlinks,citecolor=citecolor,linkcolor=linkcolor]{hyperref}
\usepackage{url}

\usepackage{enumerate}
\usepackage{tikz}
\usetikzlibrary{tikzmark}
\usepackage{xfrac}

\usepackage{makecell}
\usepackage{afterpage}

\usepackage[ruled, vlined]{algorithm2e}

\usepackage{longtable}
\usepackage{caption}
\usepackage{threeparttable}
\newcolumntype{L}[1]{>{\raggedright\let\newline\\\arraybackslash\hspace{0pt}}m{#1}}
\newcolumntype{C}[1]{>{\centering\let\newline\\\arraybackslash\hspace{0pt}}m{#1}}
\newcolumntype{R}[1]{>{\raggedleft\let\newline\\\arraybackslash\hspace{0pt}}m{#1}}

\newlength\savewidth

\newcommand{\ignorethis}[1]{}

\makeatletter
\DeclareRobustCommand\onedot{\futurelet\@let@token\@onedot}
\def\@onedot{\ifx\@let@token.\else.\null\fi\xspace}

\makeatother

\makeatletter
\def\adl@drawiv#1#2#3{%
        \hskip.5\tabcolsep
        \xleaders#3{#2.5\@tempdimb #1{1}#2.5\@tempdimb}%
                #2\z@ plus1fil minus1fil\relax
        \hskip.5\tabcolsep}
\newcommand{\cdashlinelr}[1]{%
  \noalign{\vskip\aboverulesep
           \global\let\@dashdrawstore\adl@draw
           \global\let\adl@draw\adl@drawiv}
  \cdashline{#1}
  \noalign{\global\let\adl@draw\@dashdrawstore
           \vskip\belowrulesep}}
\makeatother

\definecolor{citecolor}{HTML}{0071bc}
\definecolor{mydarkblue}{rgb}{0,0.08,1}
\definecolor{mydarkgreen}{rgb}{0.02,0.6,0.02}
\definecolor{mydarkred}{rgb}{0.8,0.02,0.02}
\definecolor{mydarkorange}{rgb}{0.40,0.2,0.02}
\definecolor{mypurple}{RGB}{111,0,255}
\definecolor{myred}{rgb}{1.0,0.0,0.0}
\definecolor{mygold}{rgb}{0.75,0.6,0.12}
\definecolor{mydarkgray}{rgb}{0.66, 0.66, 0.66}

\definecolor{darkblue}{rgb}{0,0.08,1}
\definecolor{darkgreen}{rgb}{0.02,0.6,0.02}
\definecolor{darkred}{rgb}{0.8,0.02,0.02}
\definecolor{darkorange}{rgb}{0.40,0.2,0.02}
\definecolor{darkpurple}{RGB}{111,0,255}
\definecolor{haishengcolor}{RGB}{184, 134, 11}

\definecolor{citecolor}{HTML}{0071BC}
\definecolor{linkcolor}{HTML}{ED1C24}

\definecolor{mydarkblue}{rgb}{0,0.08,1}

\theoremstyle{definition}

\newif\ifarxiv

\title{\textbf{\method}: \\ Flash Vision-Language-Action Inference for \\ Autonomous Driving}

\author{
\textbf{Zekai Li}\textsuperscript{1\,$*$} \quad
\textbf{Yihao Liang}\textsuperscript{2\,$*$} \quad
\textbf{Hongfei Zhang}\textsuperscript{3} \quad
\textbf{Jian Chen}\textsuperscript{1} \quad \\
\textbf{Yesheng Liang}\textsuperscript{1} \quad
\textbf{Zhijian Liu}\textsuperscript{1}\\
$^1$UC San Diego \quad $^2$Princeton \quad $^3$Independent Researcher\\
\textsuperscript{$*$}Indicates equal contributions \\
\\ \url{https://z-lab.ai/projects/flashdrive}
}

\iclrfinalcopy

\begin{document}

\maketitle

\begin{abstract}

Vision-Language-Action (VLA) models promise to bring end-to-end reasoning to autonomous driving, but their computational cost remains far too high for real-time control. The core challenge is structural: VLA inference is not a single bottleneck but a cascade of four. Visual encoding wastes compute on overlapping video frames; language-model prefill recomputes context that could be carried over from the previous timestep; reasoning tokens are generated serially despite low entropy; and flow-matching denoising applies uniform compute to a non-uniform velocity field. Addressing any one stage in isolation leaves the others untouched.
We propose \method\footnote[2]{Code and pre-trained checkpoints are available at \url{https://github.com/z-lab/flashdrive}.}, an algorithm-system co-design framework that targets all four stages simultaneously. Our key insight is that each bottleneck admits a distinct, lightweight algorithmic shortcut: temporal overlap enables streaming KV-cache reuse across frames; the low per-token entropy and strong intra-block correlations of driving-domain reasoning make a non-autoregressive diffusion drafter highly effective for speculative decoding; and the velocity field's structure---sharp at the endpoints, flat in the middle---permits adaptive step caching that concentrates compute where it matters. Layered on system-level CUDA Graph compilation and kernel fusion, these techniques compound. Applied to Alpamayo 1.5-10B with W4A8 quantization, \method reduces end-to-end latency from 717~ms to 151~ms ($4.7\times$) while leaving accuracy essentially unchanged: $\text{minADE}_{6}$@6.4s shifts by only $\sim$0.08~m, $\text{minADE}_{1}$ improves, and closed-loop collision and off-road rates improve in simulation. By raising a 10B-parameter reasoning VLA from 1.4~Hz to 6.6~Hz on a single GPU, \method moves end-to-end autonomous driving substantially closer to real-time deployment.

\end{abstract}

\begin{figure}[htb]
    \centering
    \includegraphics[width=\textwidth]{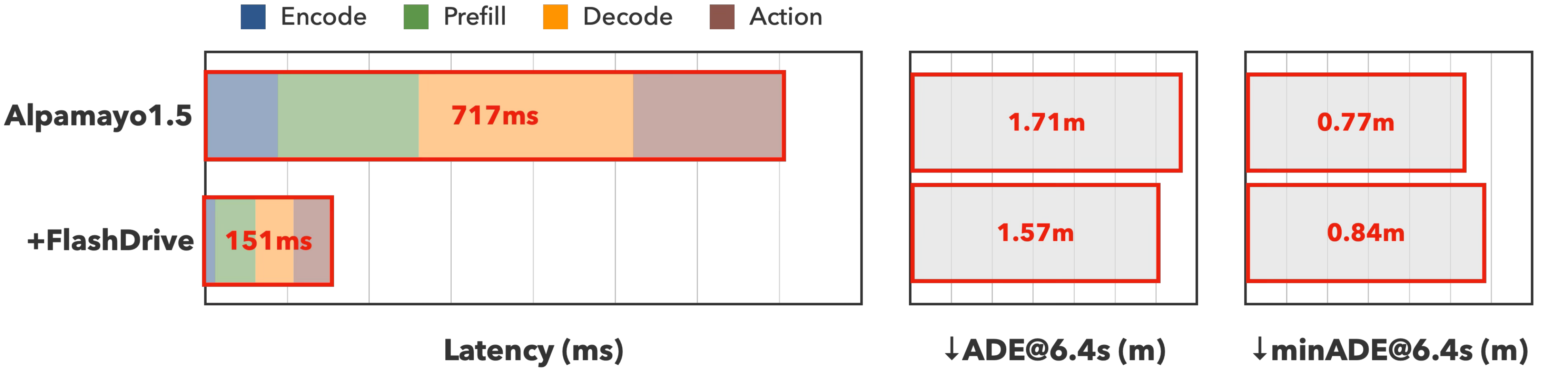}
    \caption{Reasoning VLA models for autonomous driving, such as Alpamayo 1.5, exhibit prohibitive end-to-end inference latency: 717~ms on an RTX PRO 6000, far exceeding real-time requirements. \method achieves a $4.7\times$ latency reduction (down to 151~ms) while incurring negligible degradation on $\text{minADE}_{6}$@6.4s and even improving $\text{minADE}_{1}$@6.4s.}
    \label{fig:teaser}
\end{figure}

\section{Introduction}
\label{sec:intro}

Vision-Language-Action (VLA) models---networks that reason directly over raw sensor streams and predict continuous trajectories without hand-crafted inter-module interfaces---have emerged as the leading paradigm for end-to-end autonomous driving~\citep{wang2025alpamayo, zhou2025autovla, tian2024drivevlm}. By unifying visual understanding, deliberative reasoning, and trajectory prediction within a single model, they can address complex long-tail scenarios that expose the brittleness of modular pipelines. The same architectural richness that gives VLAs this generality, however, comes at a steep computational cost.

The recent open-source Alpamayo 1.5-10B model~\citep{wang2025alpamayo}, for instance, requires 717~ms per frame on an NVIDIA RTX PRO 6000 GPU (Fig.~\ref{fig:teaser}), yielding a control frequency of just 1.4~Hz, far too slow for safe driving. Critically, this bottleneck is not specific to one model; it is structural. Every reasoning VLA of this design must encode high-resolution multi-view video, attend over long token sequences, autoregressively generate reasoning chains, and iteratively denoise trajectories. Each stage is expensive for a fundamentally different reason, and optimizing one in isolation leaves the others untouched.

To map out these bottlenecks, we decompose the VLA pipeline into four stages and profile each:
\begin{itemize}[leftmargin=1.5em,itemsep=2pt,topsep=2pt]
\item \textit{Encode}: the vision tower processes multi-view, multi-frame images. In a sliding-window setup, $\sim$75\% of these frames have already been seen, yet the model re-encodes them from scratch.
\item \textit{Prefill}: the VLM ingests thousands of image and text tokens to populate the KV cache, repeating work that could be carried over from the previous timestep.
\item \textit{Decode}: the model autoregressively generates reasoning tokens one at a time, even though driving-domain reasoning chains are short and predictable.
\item \textit{Action}: a multi-step flow-matching module runs many denoising iterations, even though the velocity field is nearly constant through the middle of the denoising path and only changes meaningfully at its endpoints.
\end{itemize}

The key observation is that each stage harbors a distinct form of redundancy, and each admits a correspondingly distinct algorithmic shortcut. This motivates \method, an algorithm-system co-design framework. On the \textbf{system side}, we compile each pipeline stage into a CUDA Graph and fuse attention and MLP kernels, yielding a $1.40\times$ speedup before any algorithmic change. On the \textbf{algorithm side}, we exploit the stage-specific redundancies:

\begin{itemize}[leftmargin=1.5em,itemsep=2pt,topsep=2pt]
\item \textbf{Streaming inference} exploits the temporal overlap in driving video: we encode only the new frame and reuse the KV cache from preceding frames, cutting the encode and prefill cost by $\sim$3$\times$. A lightweight streaming fine-tuning procedure adapts the action expert to the resulting distributional shift.

\item \textbf{Speculative reasoning} exploits two properties of domain-specific reasoning: low per-token entropy (which drives high accept rates) and strong intra-block token correlations (which favor a non-autoregressive drafter). We use DFlash~\citep{chen2026dflash}, a diffusion-based parallel drafter, to generate entire candidate blocks at once. A two-layer draft model achieves an average accepted length of $\sim$5.6 tokens, delivering a $4.7\times$ decoding speedup over the unoptimized baseline ($2.9\times$ over the system-optimized baseline).

\item \textbf{Adaptive-step flow matching} exploits the non-uniform structure of the denoising velocity field: velocities change sharply at the endpoints, where the trajectory departs from noise and snaps onto the data manifold, but are nearly constant through the middle. We cache the intermediate velocities and concentrate compute on the steps that matter, reusing four of eight diffusion steps with negligible accuracy loss.
\end{itemize}

We instantiate \method on Alpamayo 1.5-10B and, together with W4A8 quantization, achieve a $4.7\times$ end-to-end speedup (717~ms $\to$ 151~ms, Fig.~\ref{fig:teaser}). Crucially, the acceleration is nearly lossless: $\text{minADE}_{6}$@6.4s degrades by only 0.08~m, while $\text{minADE}_{1}$@6.4s improves by 0.13~m, likely because streaming fine-tuning acts as a regularizer that reduces prediction variance. The underlying methodology---profile each inference stage, identify its dominant redundancy, and design the matching shortcut---applies to any VLA deployment where latency is the binding constraint.

\section{Related Work}

\paragraph{VLA Models for Autonomous Driving.}
End-to-end autonomous driving has evolved rapidly from perception-only models to Vision-Language-Action (VLA) systems that close the loop from sensors to control. Early work unifies perception and planning through textual waypoints or meta-actions~\citep{cui2025drivemlm, sima2024drivelm, hwang2025emma}, while more recent models attach specialized action heads that output continuous trajectories~\citep{shao2024lmdrive, xu2024drivegpt4, fu2025orion, zhou2025autovla}. A recurring tension in this line of work is between the richness of deliberative reasoning and the strict latency constraints of real-time control; some systems resolve this by decoupling a slow VLM reasoner from a fast downstream planner~\citep{tian2024drivevlm, pan2024vlp}. Alpamayo 1.5~\citep{wang2025alpamayo} avoids this decoupling by bridging chain-of-causation reasoning with flow-matching action prediction in a single model, achieving strong accuracy but at a high computational cost. Our work takes this cost as the starting point.

\paragraph{Efficient VLA Inference.}
A growing body of work aims to make VLA models deployable on edge hardware. Architectural approaches reduce the cost of individual components: linear-time attention and KV caching~\citep{leal2024sarart, xu2025kv} tame the quadratic prefill, state-space models replace transformers entirely~\citep{liu2024robomamba}, and parallel or diffusion-based decoders bypass autoregressive generation~\citep{kim2025oft, song2025pdvla, black2025pi05}. Orthogonally, compression techniques shrink the model footprint, including lightweight backbones~\citep{wen2025tinyvla, budzianowski2025edgevla}, mixture-of-experts routing~\citep{song2024germ}, hierarchical reasoning-execution decoupling~\citep{zhang2024hirt}, layer pruning~\citep{zhang2026molevla, yue2024deervla}, extreme quantization~\citep{kim2024openvla, wang2025bitvla}, and token-level optimizations such as faster tokenization~\citep{pertsch2025fast}, token pruning~\citep{tan2025thinktwice}, and visual feature caching~\citep{xu2025vlacache}. Persistent KV-cache management across inference steps, developed for streaming autoregressive text generation~\citep{xiao2024streamingllm}, provides the conceptual basis for our cross-frame reuse design, which we extend to the multimodal setting with a streaming attention mask, pre-RoPE key storage, and a targeted fine-tuning procedure to compensate for distributional shift. However, most prior VLA efficiency methods target a single stage or a single axis of efficiency. \method is complementary: rather than redesigning the VLA architecture, it accelerates the \emph{full} inference pipeline of an existing model through co-design across all four stages, yielding compounding gains that no single-stage optimization can achieve alone.

\section{Method}
\label{sec:method}

Our central thesis is that VLA inference latency is not a single bottleneck but a cascade of four distinct ones (encode, prefill, decode, and action), each requiring a different solution. We first introduce stage-specific algorithmic techniques (\S\ref{sec:streaming}--\S\ref{sec:quant}) that exploit the unique redundancy structure of each stage, then describe system-level optimizations (\S\ref{sec:system}) that reduce execution overhead across the board. While we use Alpamayo 1.5~\citep{wang2025alpamayo} as the concrete instantiation, the techniques transfer to any VLA model with the same pipeline structure.

\subsection{Encode \& Prefill: Streaming Inference}
\label{sec:streaming}

\paragraph{Observation.} In continuous driving, most of each window's computation repeats the previous timestep's. The VLA model processes a sliding window of temporal frames (typically 4 frames $\times$ 4 views) that advances by one frame per timestep, so three out of four frames were already encoded at the previous step. Re-encoding them from scratch wastes $\sim$75\% of the visual computation and, since the prefill stage operates on these same tokens, a comparable fraction of the prefill as well. This redundancy is intrinsic to streaming deployment: unlike chat-based VLMs, where each query is independent, a driving VLA consumes a continuous, highly overlapping sensory stream.

\begin{figure}
    \centering
    \includegraphics[width=\textwidth]{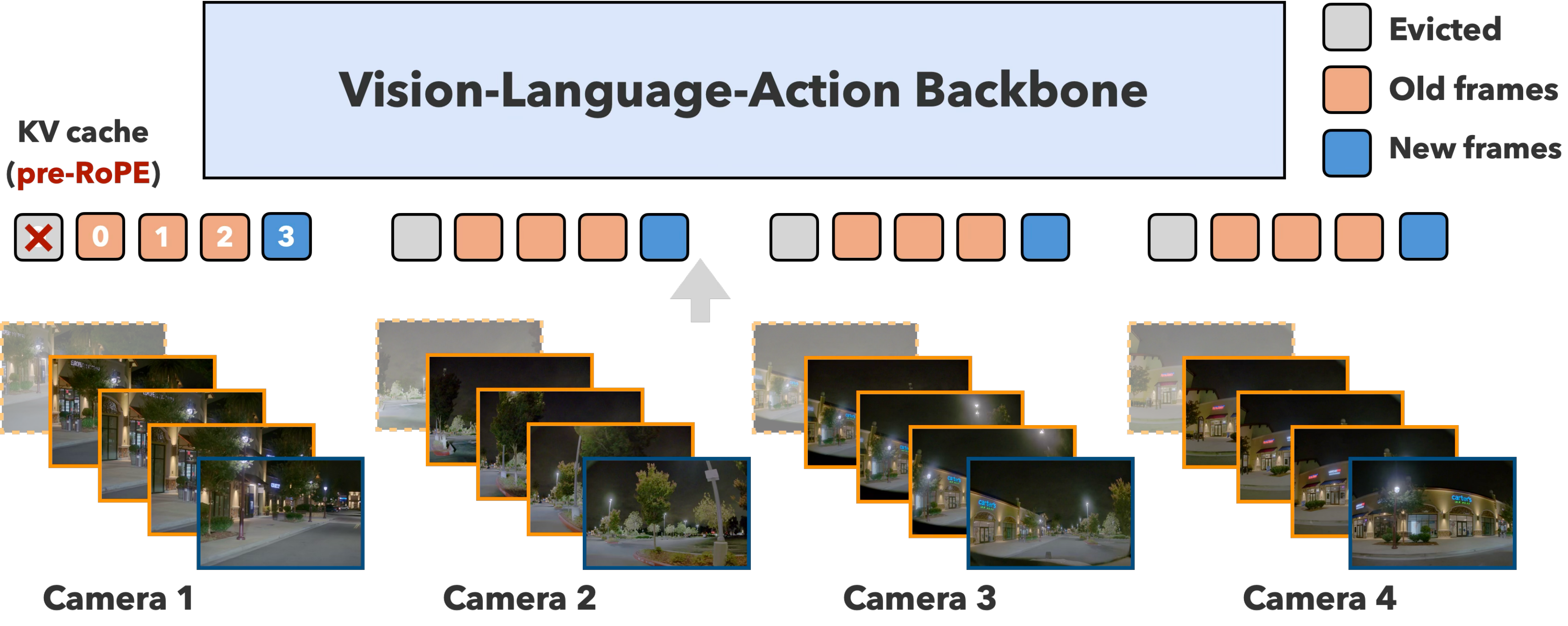}
    \caption{Streaming inference encodes only the newest frame and reuses the KV cache from preceding frames, cutting the effective sequence length by 75\%. To preserve view-major token ordering, incoming frames are inserted at the end of each camera view; to accommodate position shifts, keys are cached pre-RoPE and rotary embeddings applied on the fly.}
    \label{fig:streaming_model}
\end{figure}

\begin{figure}[htb]
    \centering
     \begin{subfigure}[b]{0.6\textwidth}
         \centering
         \includegraphics[width=\textwidth]{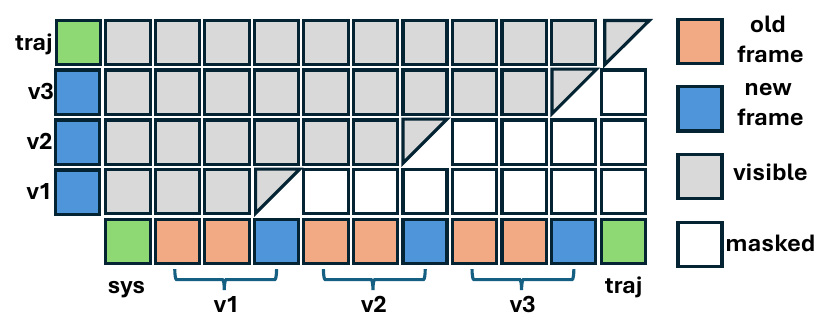}
         \caption{Streaming attention mask.}
         \label{fig:streaming_mask}
     \end{subfigure}
     \hfill
     \begin{subfigure}[b]{0.38\textwidth}
         \centering
         \includegraphics[width=\textwidth]{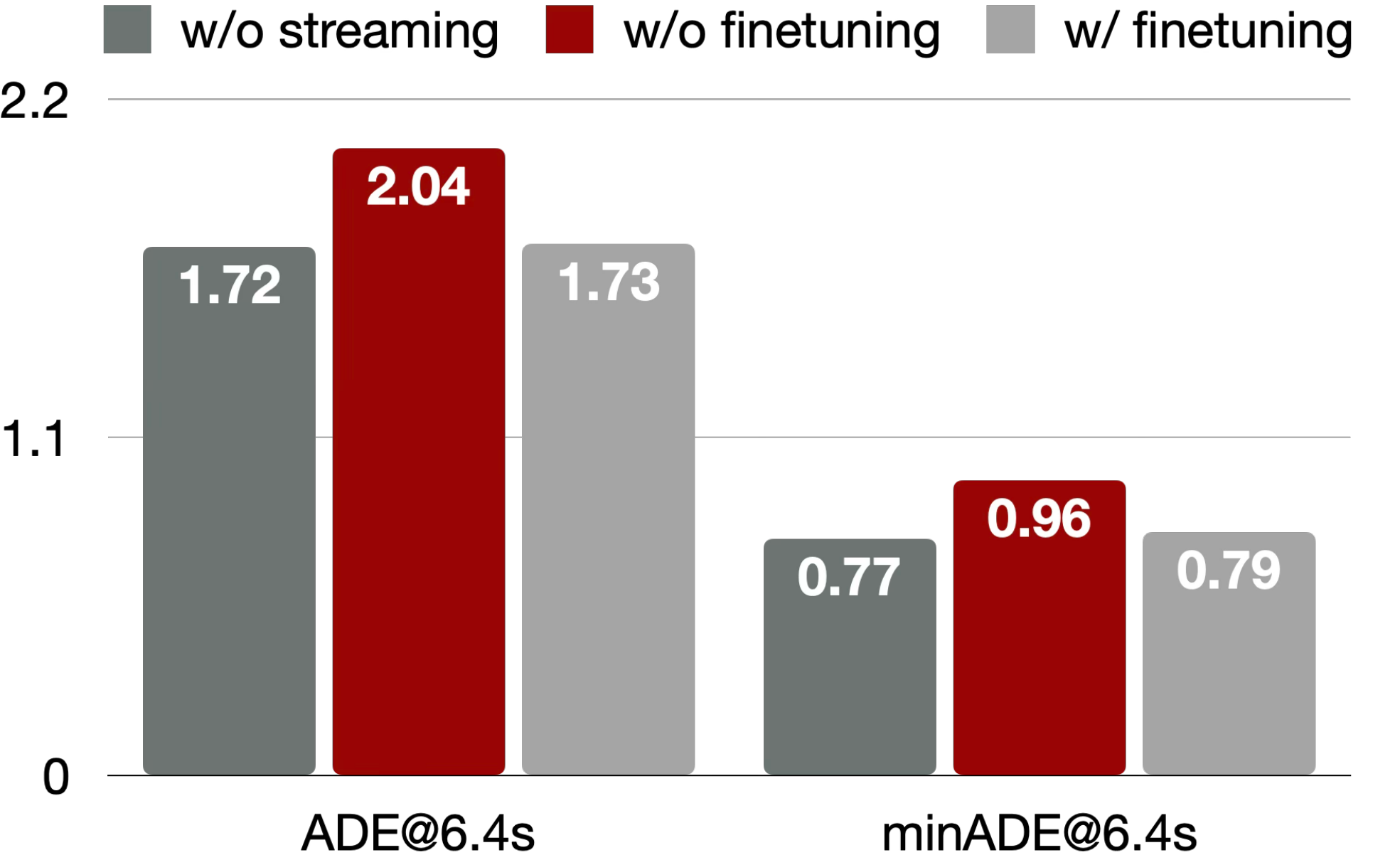}
         \caption{Accuracy with streaming fine-tuning.}
         \label{fig:streaming_finetune}
     \end{subfigure}
     \caption{(a) The streaming attention mask preserves causal attention across views while admitting only the newest frame's tokens as queries. (b) Streaming fine-tuning of the action expert recovers nearly all of the trajectory accuracy lost to KV-cache approximation.}
     \label{fig:streaming}
\end{figure}

\paragraph{Streaming design.}
We propose streaming inference (Fig.~\ref{fig:streaming_model}): encode only the newest frame and persist the KV cache from preceding frames. Two challenges arise. First, when the VLA model arranges image tokens in view-major order---all frames of one camera view laid out before the next, as in Alpamayo 1.5---simply appending new tokens would break the expected layout. Instead, we insert the new frame's tokens at the last-frame position of each view and apply a streaming attention mask (Fig.~\ref{fig:streaming_mask}) that preserves cross-view causality. Second, because visual token positions shift with each new frame, the standard post-RoPE key cache becomes stale: RoPE encodes absolute position, so a token that was at position $p$ in the previous window must now be treated as if at position $p - \Delta$. We therefore store keys pre-RoPE and apply rotary embeddings on the fly at the shifted positions. This reduces the effective sequence length by 75\%, yielding over $3\times$ speedups in both encode and prefill.

\paragraph{Streaming fine-tuning.}
The streaming KV cache is an approximation: the cached keys and values were computed under a different attention context than the current frame would produce in a full forward pass. In practice, this distributional shift degrades action accuracy by $\sim$0.3\,m $\text{minADE}_{1}$ and $\sim$0.2\,m $\text{minADE}_{6}$ (Fig.~\ref{fig:streaming_finetune}). Notably, the shift primarily affects the action expert rather than the language head. The two components use context differently: reasoning tokens are generated autoregressively and attend mainly to recent tokens, making them robust to stale cache entries from older frames. The action expert, by contrast, integrates the \emph{entire} KV cache through cross-attention to produce continuous trajectories, amplifying even small distributional mismatches.

This asymmetry suggests a targeted fix: freeze the VLM backbone and fine-tune only the action expert. Our rollout-based teacher-forcing scheme (each rollout step is driven by ground-truth frames rather than model predictions) exposes the action expert to the compounding approximation errors it will encounter at deployment: for a randomly sampled sliding window of length $L$, the model rolls out $L{-}1$ steps under the streaming mask to populate the KV cache (no gradients), then enables gradients at the final step to compute the action loss. Training on windows of varying length teaches the action expert to produce accurate trajectories even after multiple rounds of accumulated streaming approximation. This lightweight procedure recovers accuracy to near-baseline levels (1.73\,m $\text{minADE}_{1}$ and 0.79\,m $\text{minADE}_{6}$ in Fig.~\ref{fig:streaming_finetune}). We provide an additional comparison with fine-tuning only the VLM in \S\ref{sec:finetune_vlm}.

\subsection{Decode: Speculative Reasoning}
\label{sec:decode}

\paragraph{Observation.} Autoregressive reasoning is the single largest latency contributor: in our profiling, decoding accounts for 271.7~ms, or 37.9\% of the total latency, at just 56.4 tokens-per-second throughput. Recent driving VLAs generate explicit reasoning tokens to guide trajectory prediction~\citep{zhou2025autovla, wang2025alpamayo}; in Alpamayo 1.5, these chain-of-causation (CoC) tokens describe the ego vehicle state, nearby obstacles, and intended maneuvers in a structured, template-like format. Because decoding lies on the critical path between perception and action, every extra reasoning token directly delays the downstream trajectory head and therefore the control loop. The question is whether this cost is intrinsic to reasoning or an artifact of token-by-token generation.

\begin{wrapfigure}{r}{0.55\textwidth}
    \includegraphics[width=0.55\textwidth]{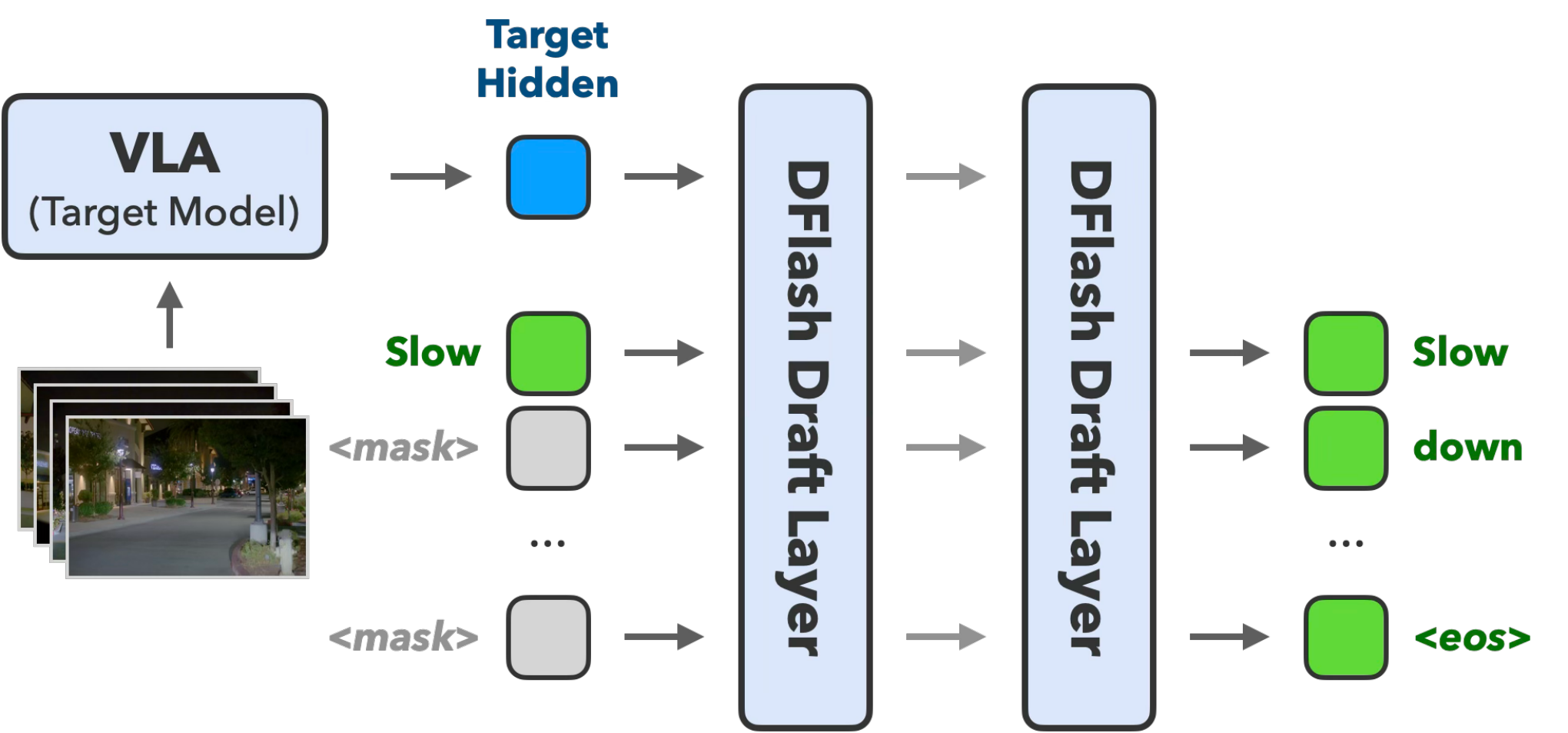}
    \caption{Speculative reasoning with DFlash. A diffusion drafter generates an entire candidate block in one forward pass, conditioned on the hidden states of the last eight target-model tokens. With only two draft layers, it achieves an average accepted length of 5.6 tokens.}
    \label{fig:dflash}
\end{wrapfigure}

\paragraph{Why driving-domain reasoning is easy to draft.}
We argue that it is largely the latter. Reasoning sequences are short ($\sim$16 tokens), follow a highly structured template, and are conditioned on a rich visual context that already determines most of the content. Unlike open-ended dialogue, the driving scene constrains the candidate space through lane topology, nearby agents, and the ego vehicle's current motion, so many plausible reasoning traces share the same prefix. This makes the per-token entropy substantially lower than in open-ended language generation, creating an opportunity for speculative decoding with high acceptance rates. Moreover, the structured nature of CoC reasoning means that tokens within a block are strongly correlated (e.g., a lane-change decision constrains the subsequent speed and trajectory descriptions), which favors drafters that generate entire blocks at once rather than one token at a time.

\paragraph{Diffusion-based drafting.}
We adopt DFlash~\citep{chen2026dflash}, which uses a diffusion language model as a non-autoregressive parallel drafter (Fig.~\ref{fig:dflash}). Unlike conventional sequential drafters, the diffusion model produces an entire block of candidates in a single forward pass, naturally capturing the intra-block correlations of structured reasoning. We train a lightweight two-layer draft model (block size\,=\,8) on $\sim$60k clips from the NVIDIA Autonomous Vehicle Dataset~\citep{nvidia2025physicalai}. Instead of using all hidden states as DFlash does, which incurs much higher verification cost, we fuse only the hidden states of the last eight target-model tokens into the draft layer's KV cache; these tokens already encode the driving context needed to predict the next reasoning block. This configuration achieves an average accepted length of 5.6 tokens, reducing decoding latency to 58.2~ms, a $4.7\times$ speedup over the unoptimized baseline (equivalently, $2.9\times$ over the system-optimized baseline used as the ablation reference in \S\ref{sec:evaluation}).

\begin{figure}[t]
    \centering
     \begin{subfigure}[b]{0.48\textwidth}
         \centering
         \includegraphics[width=\textwidth]{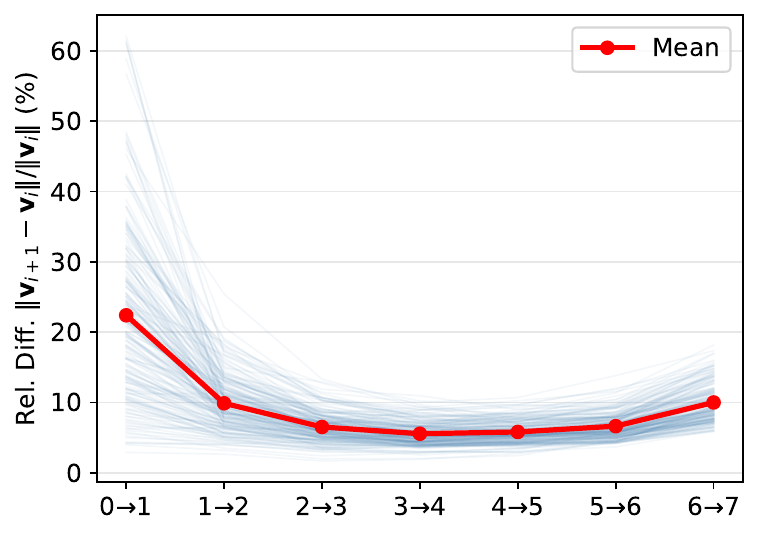}
         \caption{The normalized relative differences between velocities from consecutive steps are high at the beginning and end, but low in the middle (U-shape).}
         \label{fig:action_cache_diff}
     \end{subfigure}
     \hfill
     \begin{subfigure}[b]{0.48\textwidth}
         \centering
         \includegraphics[width=\textwidth]{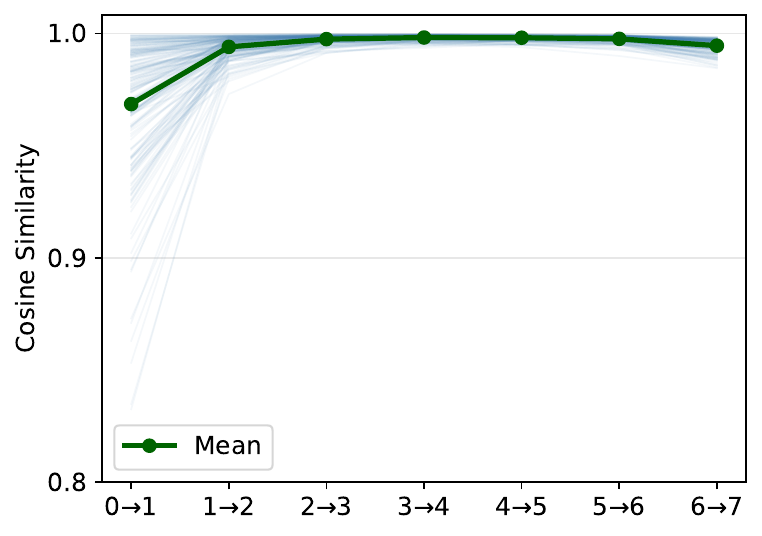}
         \caption{The cosine similarity between velocities from consecutive steps is low at the beginning and end, but high in the middle (inverted U-shape).}
         \label{fig:action_cache_cosin}
     \end{subfigure}
     \caption{Consecutive velocities in the flow-matching process are highly redundant through the middle of the denoising path but change sharply at the endpoints, so intermediate evaluations can be cached and skipped. We randomly sample 10 clips and 20 window inputs per clip to plot the relative difference (a) and cosine similarity (b) between consecutive steps.}
     \label{fig:action_cache}
\end{figure}

\subsection{Action: Adaptive-Step Flow Matching}
\label{sec:action}

\paragraph{Observation.} The flow-matching action head is one of the dominant latency contributors. To bridge language-level reasoning and continuous vehicle control, the head converts the VLM's hidden representations into trajectory waypoints through iterative denoising; in our evaluated 8-step solver, each step requires a full forward pass through the action network.

\paragraph{Why uniform step reduction fails.}
The na\"ive remedy---fewer uniformly spaced steps---treats all regions of the denoising trajectory as equally important, an assumption that turns out to be wrong. Profiling the velocity field $v_t$ across the denoising trajectory (Fig.~\ref{fig:action_cache}) reveals a striking structure: the normalized relative differences between consecutive velocities trace a U-shape (Fig.~\ref{fig:action_cache_diff}), and their cosine similarities the corresponding inverted U (Fig.~\ref{fig:action_cache_cosin}). The velocity changes sharply at the first and last steps, where the trajectory departs from the noise prior and converges onto the data manifold, but is nearly constant through the middle, where the ODE flow traverses a smooth, low-curvature region of the learned vector field.

\paragraph{Insight.} This non-uniformity has a clear physical interpretation: the early steps establish the coarse trajectory structure (lane choice, turn direction), the final steps snap the prediction onto the manifold of physically plausible trajectories (satisfying kinematic constraints and road geometry), and the intermediate steps perform only minor refinements to an already well-determined path. The endpoints carry the signal; the middle carries the inertia.

\paragraph{Adaptive caching.}
We exploit this structure by caching the velocity at the middle steps and reusing it in lieu of recomputation. Concretely, after evaluating the first few and last few steps with fresh network calls, we replace four intermediate evaluations with the cached velocity from the preceding step. This cuts action latency from 113.9\,ms to 47.6\,ms while concentrating compute on the steps that shape the trajectory the most. As shown in Tab.~\ref{tab:speedup_breakdown}, the adaptive strategy is near-lossless: $\text{minADE}_{6}$ increases by only 0.04\,m, while $\text{minADE}_{1}$ improves by 0.14\,m---skipping redundant mid-trajectory refinements appears to reduce accumulated numerical error from the ODE solver.

\subsection{Quantization}
\label{sec:quant}

Large VLA models often exceed the memory capacity of consumer-grade GPUs: Alpamayo 1.5-10B requires $\sim$31.6\,GB in FP16 to generate 6 trajectory samples. Quantization is the standard remedy, but it presents a choice. Methods like AWQ~\citep{lin2024awq} quantize only the weights to 4-bit (W4A16): this helps memory-bound decoding by shrinking the data the GPU must load per token, but leaves the compute-bound prefill stage untouched. For a chatbot LLM where decoding dominates, that trade-off is acceptable. For a VLA model with thousands of vision tokens in every prompt, prefill is too expensive to ignore.

\textbf{W4A8} quantization targets both regimes: 4-bit weights cut memory bandwidth for decoding, while 8-bit activations unlock faster INT8 matrix multiplies for the compute-heavy prefill. We apply ParoQuant~\citep{liang2026paroquant} to quantize the VLM backbone's weights to 4-bit and execute inference with 8-bit activations using the W4A8 variant of the Marlin kernels~\citep{frantar2024marlin}; the action expert, whose continuous outputs are most sensitive to numerical error, remains in BF16. This cuts the memory footprint to $\sim$18.3\,GB while reducing the remaining end-to-end latency by a further 14\% (176.0\,ms $\to$ 151.4\,ms).

\subsection{System Optimizations}
\label{sec:system}

The algorithmic techniques above reduce the amount of computation; system-level optimizations reduce the cost of executing what remains. Unlike standard LLM serving, where the workload is dominated by a single homogeneous decode loop, VLA inference chains together four heterogeneous stages (vision encoder, language prefill, autoregressive decode, flow-matching action), each with its own kernel mix. This heterogeneity amplifies the CPU-side dispatch cost: the pipeline involves hundreds of small kernel launches per forward pass, and at the low arithmetic intensities typical of single-batch decoding, launch overhead becomes a significant fraction of wall-clock time.

\paragraph{CUDA Graph.}
We compile each pipeline stage into a CUDA Graph, which records the full kernel sequence and replays it in a single GPU-side launch. This is particularly impactful for the decode stage, where the GPU would otherwise idle between every token while waiting for the CPU to schedule the next kernel.

\paragraph{Kernel Fusion.}
Many VLM backbones dispatch the Q, K, and V projections as three separate kernels, and likewise for the gate and up projections in the MLP, totaling six launches where two suffice. We fuse each group and additionally compile under the max-autotune mode, which merges consecutive element-wise and reduction operations and auto-selects the fastest implementation. Together, CUDA Graphs and kernel fusion provide a $1.40\times$ speedup without changing the model's computation.

\section{Experiments}
\label{sec:evaluation}

\subsection{Setup}

\paragraph{Model.} We evaluate \method on Alpamayo 1.5-10B, a recent open-source state-of-the-art VLA model for autonomous driving. While we use Alpamayo 1.5-10B as the primary testbed, \method's techniques are applicable to any VLA model sharing the encode--prefill--decode--action pipeline. Results of \method on Alpamayo 1 (released as Alpamayo-R1) are provided in Appendix~\ref{sec:apmy_1}.

\paragraph{Dataset.} We use the open-source NVIDIA Autonomous Vehicle Dataset~\citep{nvidia2025physicalai} for both training and evaluation. For streaming fine-tuning, we sample 4k clips across different chunks and randomly select starting points and rollout lengths within each clip, yielding $\sim$600k training samples (at most 150 per clip). For diffusion draft-model training, we randomly sample one window input from each of 60k clips in the dataset. For evaluation, we sample 100 clips and extract sliding-window inputs at 10\,FPS, producing 120 windows per clip and 12k evaluation samples in total.

\paragraph{Metrics.} Following the official protocol, we report $\text{minADE}_{6}$@6.4s, the minimum L2 distance between the ground-truth trajectory and each of six predicted trajectories over the next 6.4\,s. We additionally report $\text{minADE}_{1}$@6.4s, the ADE of a single predicted trajectory over the same horizon.

\subsection{Efficiency Results}

\begin{table}[t]
\centering
\resizebox{\textwidth}{!}{%
\small
\begin{tabular}{llllllll}
\toprule
 & \multicolumn{5}{c}{Latency (with 1 Trajectory) (ms)} & \multicolumn{2}{c}{Trajectory Error (m)} \\
\cmidrule(lr){2-6}\cmidrule(lr){7-8}
 & Encode & Prefill & Decode & Action & Total & minADE\textsubscript{1} $\downarrow$ & minADE\textsubscript{6} $\downarrow$ \\
\midrule
Alpamayo 1.5 & 87.0 & 165.3 & 271.7 \trpt{56.4} & 192.9 & 716.9 & 1.705 & 0.767 \\
\midrule
+ System Optimizations               & 40.5 & 188.3 & 170.6 \trpt{89.5} & 113.9 & 513.3 & 1.673 & 0.770 \\
\, \,+ Streaming Inference       & 12.0 & 59.6  & 170.5 \trpt{89.5} & 113.8 & 355.8 & 1.697 & 0.791 \\
\, \,+ Speculative Reasoning               & 40.7 & 187.3 & 58.2 \trpt{254.4} & 115.0 & 401.3 & 1.633 & 0.777 \\
\, \,+ Adaptive-Step Flow Matching          & 40.9 & 188.3 & 173.6 \trpt{89.1} & 47.6  & 450.4 & 1.536 & 0.812 \\
\, \,+ All above            & 12.1 & 58.9 & 57.5 \trpt{255.6} & 47.5  & 176.0 & 1.563 &  0.850 \\
\, \, \, \,+ Quantization    & 12.0 & 47.2 & 45.3 \trpt{326.9} & 46.9  & 151.4 & 1.573 & 0.844 \\
\bottomrule
\end{tabular}%
}
\caption{\method achieves a 4.7$\times$ overall speedup with one trajectory on an NVIDIA RTX PRO 6000, while minADE\textsubscript{1} improves and minADE\textsubscript{6} degrades by only $\sim$0.08\,m. Each algorithmic technique is first ablated individually on top of the system-optimized baseline, then combined (``All above'') and quantized. The \trpt{throughput} next to the decoding time denotes token throughput (tokens/s) for that setting.}
\label{tab:speedup_breakdown}
\end{table}

\paragraph{End-to-end speedup.}
As shown in Tab.~\ref{tab:speedup_breakdown}, \method achieves a $4.7\times$ end-to-end speedup, reducing latency to 151.4~ms and raising the control frequency from 1.4\,Hz to 6.6\,Hz---a qualitative change in deployability, moving VLA inference from clearly unusable to within the replanning rates of urban driving.

\paragraph{System-level gains.}
CUDA Graphs and kernel fusion alone reduce end-to-end latency by 28.4\% ($1.40\times$). The gains are not uniform across stages: encode, decode, and action benefit the most because their many small kernels make them disproportionately sensitive to launch overhead. Prefill, dominated by a single long matrix multiplication, sees no meaningful improvement; the 23\,ms increase relative to the raw baseline lies within typical measurement variance for single-batch runs. This asymmetry underscores why system and algorithm optimizations are complementary: system-level changes address overhead, while algorithmic changes address the computation itself.

\paragraph{Algorithm-level gains.}
Tab.~\ref{tab:speedup_breakdown} ablates each technique on top of the system-optimized baseline. Streaming inference accelerates encoding by $3.4\times$ and prefilling by $3.2\times$. DFlash reduces decoding latency by $2.9\times$ over the system-optimized baseline ($4.7\times$ over the unoptimized baseline), directly targeting the single largest latency contributor. Adaptive-step flow matching reuses the velocity on four of eight diffusion steps, yielding a $2.4\times$ action-stage speedup. Jointly, the algorithmic gains save 337.3~ms, roughly $1.7\times$ the system-level savings, confirming that eliminating redundant computation matters more than eliminating overhead when both are present. W4A8 quantization contributes a further 24.6~ms reduction, and its distribution across stages validates the choice of 8-bit activations: decode benefits from the smaller weights (lower memory-bandwidth pressure), while prefill benefits from INT8 tensor cores---a gain that weight-only W4A16 quantization would forgo entirely.

\paragraph{Cross-device deployment.}
Beyond the RTX PRO 6000, we benchmark \method on the Jetson Thor, RTX~3090, RTX~4090, and RTX~5090 (Tab.~\ref{tab:devices}). With a single trajectory sample, \method achieves consistent speedups ranging from $4.0\times$ to $6.0\times$, and notably enables VLA deployment on edge devices: the Jetson Thor sees a $4.0\times$ speedup.

The advantage becomes more pronounced with six trajectory samples: \method reaches $9.6\times$ on Jetson Thor, $10.0\times$ on RTX~5090, and $10.6\times$ on RTX PRO 6000. On the RTX~3090 and RTX~4090, the unoptimized Alpamayo 1.5 model fails due to their 24\,GB VRAM limit, while \method still runs at 694.8\,ms and 404.1\,ms, respectively. These results highlight \method's deployment readiness, particularly given that real-world autonomous driving typically requires multiple trajectory samples for downstream planning.

\begin{table}[htb]
\small
\centering
\renewcommand{\arraystretch}{1.2}
\resizebox{\textwidth}{!}{
    \begin{tabular}{llccccc}
    \toprule
    \multicolumn{2}{c}{Model}  & Jetson Thor  & RTX 3090  & RTX 4090  & RTX 5090  & RTX PRO 6000  \\ \midrule
    \multirow{2}{*}{1 sample}  & Alpamayo1.5  & 3770.3   & 1891.9     & 1307.1    & 878.1   & 716.9   \\
    & \method & 943.6 (\textcolor{darkgreen}{4.0$\times$}) & 382.3 (\textcolor{darkgreen}{4.9$\times$}) & 217.2 (\textcolor{darkgreen}{6.0$\times$}) & 183.7 (\textcolor{darkgreen}{4.8$\times$}) & 151.4 (\textcolor{darkgreen}{4.7$\times$}) \\ \midrule
    \multirow{2}{*}{6 samples} & Alpamayo1.5   &  14596.5   & OOM$^\dagger$   & OOM$^\dagger$    & 3163.5   & 2609.8 \\
    & \method & 1522.6 (\textcolor{darkgreen}{9.6$\times$})  & 694.8$^\dagger$   & 404.1$^\dagger$   & 317.9 (\textcolor{darkgreen}{10.0$\times$})  & 245.8 (\textcolor{darkgreen}{10.6$\times$})  \\ \bottomrule
    \end{tabular}
}
\caption{FlashDrive demonstrates robust acceleration across five GPUs spanning edge, consumer, and workstation classes, achieving up to a $6.0\times$ speedup with one trajectory sample on an RTX~4090. \method's advantage is more pronounced with six samples, reaching $10.0\times$ on an RTX~5090 and $10.6\times$ on an RTX PRO 6000. $^\dagger$No speedup ratio is shown where the baseline OOM'd; \method latencies are reported for reference.}
\label{tab:devices}
\end{table}

\subsection{Open-loop Accuracy}

\method's acceleration is nearly lossless in open-loop accuracy. Tab.~\ref{tab:speedup_breakdown} compares \method against the Alpamayo 1.5 baseline under single-trajectory $\text{minADE}_{1}$@6.4s and six-sample $\text{minADE}_{6}$@6.4s. Despite combining streaming inference, speculative reasoning, adaptive-step flow matching, and W4A8 quantization, \method shifts $\text{minADE}_{6}$@6.4s only from 0.767\,m to 0.844\,m. This $\sim$0.08\,m increase corresponds to roughly 1.2\,cm per second over the 6.4\,s prediction horizon---negligible relative to lane-level planning tolerances. Meanwhile, $\text{minADE}_{1}$@6.4s \emph{improves} from 1.705\,m to 1.573\,m: the acceleration techniques do not merely trade accuracy for latency; some components stabilize the generated trajectory. The likely mechanism is that streaming fine-tuning trains the action expert to be robust to approximate KV caches, acting as a regularizer that reduces prediction variance. The ablation rows support this reading: individual algorithmic changes perturb accuracy in different directions, but their combination preserves the overall driving quality while delivering the full $4.7\times$ speedup.

\subsection{Closed-loop Evaluation}
\begin{table}[t]
\centering
\renewcommand{\arraystretch}{1.2}
\resizebox{\textwidth}{!}{%
\begin{tabular}{l|cccccccc}
\toprule
& D\textsubscript{traj} (m) $\downarrow$ & D\textsubscript{loc} (m) $\downarrow$ & P\textsubscript{rel} $\uparrow$ & Collision $\downarrow$ & Off Road $\downarrow$ & Wrong Lane $\downarrow$ & Plan Dev. $\downarrow$ & Latency (ms) $\downarrow$ \\ \midrule
Alpamayo 1.5 & \textbf{20.0}  & 35.2   & 0.85  & 0.19   & 0.41  & \textbf{0.45}  & 0.24 & 1150 \\
+FlashDrive  & 22.4  & \textbf{32.8}   & \textbf{0.85}  & \textbf{0.15}   & \textbf{0.32}  & 0.51  & \textbf{0.16} & \textbf{463} (\textcolor{darkgreen}{2.5$\times$})  \\  \bottomrule                                                  
\end{tabular}
}
\caption{Closed-loop evaluation using AlpaSim. FlashDrive achieves safer driving with comparable ground-truth alignment under one sampled trajectory. In addition, our $2.5\times$ per-step rollout speedup improves closed-loop training and evaluation efficiency.}
\label{tab:alpasim}
\end{table}

\method preserves closed-loop driving quality (Tab.~\ref{tab:alpasim}; metric definitions in Tab.~\ref{tab:alpasim_metrics}). We evaluate on 100 randomly sampled clips in AlpaSim~\citep{alpasim_2025}, an open-source, modular platform for closed-loop evaluation of end-to-end driving policies with realistic sensor, dynamics, and traffic simulation. On trajectory-tracking metrics, \method has a slightly larger maximum distance to the full ground-truth path ($D_{\text{traj}}$: 22.4\,m vs. 20.0\,m) but a lower time-aligned localization error ($D_{\text{loc}}$: 32.8\,m vs. 35.2\,m) and identical relative progress ($P_{\text{rel}}$: 0.85): the accelerated policy completes the same routes while tracking the rollout timing more closely.

The event-based safety metrics improve: the collision rate drops from 0.19 to 0.15 and the off-road rate from 0.41 to 0.32, indicating fewer episodes with safety-critical failures, while the plan-deviation score falls substantially (0.24 to 0.16), reflecting more temporally consistent successive plans. The one metric that regresses is Wrong Lane (0.45 to 0.51); it measures whether heading deviates from the lane centerline by more than $2\pi/3$ and is particularly sensitive near intersections and merge zones, where transient heading deviations are common. Taken together, the closed-loop results suggest that acceleration does not compromise driving safety.

Beyond driving quality, \method achieves a $2.5\times$ speedup on per-step rollout latency in AlpaSim---covering model inference, trajectory optimization, and simulator rendering---from 1150\,ms to 463\,ms. This directly improves data collection efficiency for VLA models in simulation: at the same compute budget, \method evaluates substantially more closed-loop episodes, making large-scale training, stress testing, and ablation studies more practical.

\section{Conclusion}
\label{sec:conclusion}

We presented \method, an algorithm-system co-design framework that brings VLA-based autonomous driving substantially closer to real-time operation on a single GPU, including consumer-grade and edge devices. The key lesson from our work is that VLA inference is not a monolithic bottleneck but a cascade of four stages, each dominated by a different form of redundancy: temporal overlap in vision, carry-over context in prefill, serialization in reasoning, and over-iteration in denoising. By matching each bottleneck to a lightweight algorithmic shortcut (streaming, speculative decoding, adaptive step caching) and layering them on top of system-level compilation and fusion, the speedups compound to $4.7\times$ with negligible accuracy loss. We believe this \emph{profile-then-exploit} methodology generalizes broadly: whenever an inference pipeline has structurally heterogeneous bottlenecks, the path to efficiency is not one universal technique applied everywhere, but the right lightweight shortcut matched to each stage.

\section*{Acknowledgment}

We gratefully acknowledge Yotta Labs for providing the compute resources supporting this work.

{\small
\bibliography{reference}

\begin{thebibliography}{34}
\providecommand{\natexlab}[1]{#1}
\providecommand{\url}[1]{\texttt{#1}}
\expandafter\ifx\csname urlstyle\endcsname\relax
  \providecommand{\doi}[1]{doi: #1}\else
  \providecommand{\doi}{doi: \begingroup \urlstyle{rm}\Url}\fi

\bibitem[Black et~al.(2025)Black, Brown, Darpinian, Dhabalia, Driess, Esmail, Equi, Finn, Fusai, Galliker, Ghosh, Groom, Hausman, Ichter, Jakubczak, Jones, Ke, LeBlanc, Levine, Li-Bell, Mothukuri, Nair, Pertsch, Ren, Shi, Smith, Springenberg, Stachowicz, Tanner, Vuong, Walke, Walling, Wang, Yu, and Zhilinsky]{black2025pi05}
Kevin Black, Noah Brown, James Darpinian, Karan Dhabalia, Danny Driess, Adnan Esmail, Michael Equi, Chelsea Finn, Niccolo Fusai, Manuel~Y. Galliker, Dibya Ghosh, Lachy Groom, Karol Hausman, Brian Ichter, Szymon Jakubczak, Tim Jones, Liyiming Ke, Devin LeBlanc, Sergey Levine, Adrian Li-Bell, Mohith Mothukuri, Suraj Nair, Karl Pertsch, Allen~Z. Ren, Lucy~Xiaoyang Shi, Laura Smith, Jost~Tobias Springenberg, Kyle Stachowicz, James Tanner, Quan Vuong, Homer Walke, Anna Walling, Haohuan Wang, Lili Yu, and Ury Zhilinsky.
\newblock {$\pi_{0.5}$: A Vision-Language-Action Model with Open-World Generalization}.
\newblock In \emph{Conference on Robot Learning (CoRL)}, 2025.

\bibitem[Budzianowski et~al.(2025)Budzianowski, Maa, Freed, Mo, Hsiao, Xie, M{\l}oduchowski, Tipnis, and Bolte]{budzianowski2025edgevla}
Pawe{\l} Budzianowski, Wesley Maa, Matthew Freed, Jingxiang Mo, Winston Hsiao, Aaron Xie, Tomasz M{\l}oduchowski, Viraj Tipnis, and Benjamin Bolte.
\newblock {EdgeVLA: Efficient Vision-Language-Action Models}.
\newblock \emph{arXiv preprint arXiv:2507.14049}, 2025.

\bibitem[Chen et~al.(2026)Chen, Liang, and Liu]{chen2026dflash}
Jian Chen, Yesheng Liang, and Zhijian Liu.
\newblock {DFlash: Block Diffusion for Flash Speculative Decoding}.
\newblock In \emph{International Conference on Machine Learning (ICML)}, 2026.

\bibitem[Cui et~al.(2025)Cui, Wang, Li, Xie, Zou, Deng, Luo, Lu, Zhu, and Dai]{cui2025drivemlm}
Erfei Cui, Wenhai Wang, Zhiqi Li, Jiangwei Xie, Haoming Zou, Hanming Deng, Gen Luo, Lewei Lu, Xizhou Zhu, and Jifeng Dai.
\newblock {DriveMLM: Aligning Multi-Modal Large Language Models with Behavioral Planning States for Autonomous Driving}.
\newblock \emph{Visual Intelligence}, 2025.

\bibitem[Frantar et~al.(2024)Frantar, Castro, Chen, Hoefler, and Alistarh]{frantar2024marlin}
Elias Frantar, Roberto~L Castro, Jiale Chen, Torsten Hoefler, and Dan Alistarh.
\newblock Marlin: Mixed-precision auto-regressive parallel inference on large language models.
\newblock \emph{arXiv preprint arXiv:2408.11743}, 2024.

\bibitem[Fu et~al.(2025)Fu, Zhang, Zhao, Cui, Liang, Zhang, Zhang, Xie, Wang, and Bai]{fu2025orion}
Haoyu Fu, Diankun Zhang, Zongchuang Zhao, Jianfeng Cui, Dingkang Liang, Chong Zhang, Dingyuan Zhang, Hongwei Xie, Bing Wang, and Xiang Bai.
\newblock {ORION: A Holistic End-to-End Autonomous Driving Framework by Vision-Language Instructed Action Generation}.
\newblock In \emph{IEEE/CVF International Conference on Computer Vision (ICCV)}, 2025.

\bibitem[Hwang et~al.(2025)Hwang, Xu, Lin, Hung, Ji, Choi, Huang, He, Covington, Sapp, Zhou, Guo, Anguelov, and Tan]{hwang2025emma}
Jyh-Jing Hwang, Runsheng Xu, Hubert Lin, Wei-Chih Hung, Jingwei Ji, Kristy Choi, Di~Huang, Tong He, Paul Covington, Benjamin Sapp, Yin Zhou, James Guo, Dragomir Anguelov, and Mingxing Tan.
\newblock {EMMA: End-to-End Multimodal Model for Autonomous Driving}.
\newblock \emph{Transactions on Machine Learning Research (TMLR)}, 2025.

\bibitem[Kim et~al.(2024)Kim, Pertsch, Karamcheti, Xiao, Balakrishna, Nair, Rafailov, Foster, Lam, Sanketi, Vuong, Kollar, Burchfiel, Tedrake, Sadigh, Levine, Liang, and Finn]{kim2024openvla}
Moo~Jin Kim, Karl Pertsch, Siddharth Karamcheti, Ted Xiao, Ashwin Balakrishna, Suraj Nair, Rafael Rafailov, Ethan Foster, Grace Lam, Pannag Sanketi, Quan Vuong, Thomas Kollar, Benjamin Burchfiel, Russ Tedrake, Dorsa Sadigh, Sergey Levine, Percy Liang, and Chelsea Finn.
\newblock {OpenVLA: An Open-Source Vision-Language-Action Model}.
\newblock In \emph{Conference on Robot Learning (CoRL)}, 2024.

\bibitem[Kim et~al.(2025)Kim, Finn, and Liang]{kim2025oft}
Moo~Jin Kim, Chelsea Finn, and Percy Liang.
\newblock {Fine-Tuning Vision-Language-Action Models: Optimizing Speed and Success}.
\newblock In \emph{Robotics: Science and Systems (RSS)}, 2025.

\bibitem[Leal et~al.(2024)Leal, Choromanski, Jain, Dubey, Varley, Ryoo, Lu, Liu, Sindhwani, Vuong, Sarl{\'o}s, Oslund, Hausman, and Rao]{leal2024sarart}
Isabel Leal, Krzysztof Choromanski, Deepali Jain, Kumar~Avinava Dubey, Jake Varley, Michael~S. Ryoo, Yao Lu, Frederick Liu, Vikas Sindhwani, Quan~Ho Vuong, Tam{\'a}s Sarl{\'o}s, Kenneth Oslund, Karol Hausman, and Kanishka Rao.
\newblock {SARA-RT: Scaling Up Robotics Transformers with Self-Adaptive Robust Attention}.
\newblock In \emph{IEEE International Conference on Robotics and Automation (ICRA)}, 2024.

\bibitem[Liang et~al.(2026)Liang, Chen, Zhang, Han, and Liu]{liang2026paroquant}
Yesheng Liang, Haisheng Chen, Zihan Zhang, Song Han, and Zhijian Liu.
\newblock {ParoQuant: Pairwise Rotation Quantization for Efficient Reasoning LLM Inference}.
\newblock In \emph{International Conference on Learning Representations (ICLR)}, 2026.

\bibitem[Lin et~al.(2024)Lin, Tang, Tang, Yang, Chen, Wang, Xiao, Dang, Gan, and Han]{lin2024awq}
Ji~Lin, Jiaming Tang, Haotian Tang, Shang Yang, Wei-Ming Chen, Wei-Chen Wang, Guangxuan Xiao, Xingyu Dang, Chuang Gan, and Song Han.
\newblock {AWQ: Activation-Aware Weight Quantization for On-Device LLM Compression and Acceleration}.
\newblock In \emph{Conference on Machine Learning and Systems (MLSys)}, 2024.

\bibitem[Liu et~al.(2024)Liu, Liu, Wang, An, Li, Zhou, Yang, Zhang, Guo, and Zhang]{liu2024robomamba}
Jiaming Liu, Mengzhen Liu, Zhenyu Wang, Pengju An, Xiaoqi Li, Kaichen Zhou, Senqiao Yang, Renrui Zhang, Yandong Guo, and Shanghang Zhang.
\newblock {RoboMamba: Efficient Vision-Language-Action Model for Robotic Reasoning and Manipulation}.
\newblock In \emph{Advances in Neural Information Processing Systems (NeurIPS)}, 2024.

\bibitem[{NVIDIA}(2025)]{nvidia2025physicalai}
{NVIDIA}.
\newblock {PhysicalAI Autonomous Vehicles Dataset}, 2025.
\newblock URL \url{https://huggingface.co/datasets/nvidia/PhysicalAI-Autonomous-Vehicles}.

\bibitem[NVIDIA et~al.(2025)NVIDIA, Cao, de~Lutio, Fidler, Cobo, Gojcic, Igl, Ivanovic, Karkus, Esturo, Pavone, Smith, Tanimura, Tyszkiewicz, Watson, Wu, and Zhang]{alpasim_2025}
NVIDIA, Yulong Cao, Riccardo de~Lutio, Sanja Fidler, Guillermo~Garcia Cobo, Zan Gojcic, Maximilian Igl, Boris Ivanovic, Peter Karkus, Janick~Martinez Esturo, Marco Pavone, Aaron Smith, Ellie Tanimura, Michal Tyszkiewicz, Michael Watson, Qi~Wu, and Le~Zhang.
\newblock Alpasim: A modular, lightweight, and data-driven research simulator for autonomous driving, October 2025.
\newblock URL \url{https://github.com/NVlabs/alpasim}.

\bibitem[Pan et~al.(2024)Pan, Yaman, Nesti, Mallik, Allievi, Velipasalar, and Ren]{pan2024vlp}
Chenbin Pan, Burhaneddin Yaman, Tommaso Nesti, Abhirup Mallik, Alessandro~G Allievi, Senem Velipasalar, and Liu Ren.
\newblock {VLP: Vision Language Planning for Autonomous Driving}.
\newblock In \emph{IEEE/CVF Conference on Computer Vision and Pattern Recognition (CVPR)}, 2024.

\bibitem[Pertsch et~al.(2025)Pertsch, Stachowicz, Ichter, Driess, Nair, Vuong, Mees, Finn, and Levine]{pertsch2025fast}
Karl Pertsch, Kyle Stachowicz, Brian Ichter, Danny Driess, Suraj Nair, Quan Vuong, Oier Mees, Chelsea Finn, and Sergey Levine.
\newblock {FAST: Efficient Action Tokenization for Vision-Language-Action Models}.
\newblock In \emph{Robotics: Science and Systems (RSS)}, 2025.

\bibitem[Shao et~al.(2024)Shao, Hu, Wang, Waslander, Liu, and Li]{shao2024lmdrive}
Hao Shao, Yuxuan Hu, Letian Wang, Steven~L. Waslander, Yu~Liu, and Hongsheng Li.
\newblock {LMDrive: Closed-Loop End-to-End Driving with Large Language Models}.
\newblock In \emph{IEEE/CVF Conference on Computer Vision and Pattern Recognition (CVPR)}, 2024.

\bibitem[Sima et~al.(2024)Sima, Renz, Chitta, Chen, Zhang, Xie, Bei{\ss}wenger, Luo, Geiger, and Li]{sima2024drivelm}
Chonghao Sima, Katrin Renz, Kashyap Chitta, Li~Chen, Hanxue Zhang, Chengen Xie, Jens Bei{\ss}wenger, Ping Luo, Andreas Geiger, and Hongyang Li.
\newblock {DriveLM: Driving with Graph Visual Question Answering}.
\newblock In \emph{European Conference on Computer Vision (ECCV)}, 2024.

\bibitem[Song et~al.(2024)Song, Zhao, Ding, Cui, Lyu, Fan, and Wang]{song2024germ}
Wenxuan Song, Han Zhao, Pengxiang Ding, Can Cui, Shangke Lyu, Yaning Fan, and Donglin Wang.
\newblock {GeRM: A Generalist Robotic Model with Mixture-of-Experts for Quadruped Robot}.
\newblock In \emph{IEEE/RSJ International Conference on Intelligent Robots and Systems (IROS)}, 2024.

\bibitem[Song et~al.(2025)Song, Chen, Ding, Zhao, Zhao, Zhong, Ge, Li, Wang, Ma, Wang, and Li]{song2025pdvla}
Wenxuan Song, Jiayi Chen, Pengxiang Ding, Han Zhao, Wei Zhao, Zhide Zhong, Zongyuan Ge, Zhijun Li, Donglin Wang, Jun Ma, Lujia Wang, and Haoang Li.
\newblock {PD-VLA: Accelerating Vision-Language-Action Model Integrated with Action Chunking via Parallel Decoding}.
\newblock \emph{arXiv preprint arXiv:2503.02310}, 2025.

\bibitem[Tan et~al.(2025)Tan, Yang, Ye, Zheng, Bai, Wang, Hao, and Chen]{tan2025thinktwice}
Xudong Tan, Yaoxin Yang, Peng Ye, Jialin Zheng, Bizhe Bai, Xinyi Wang, Jia Hao, and Tao Chen.
\newblock {Think Twice, Act Once: Token-Aware Compression and Action Reuse for Efficient Inference in Vision-Language-Action Models}.
\newblock \emph{arXiv preprint arXiv:2505.21200}, 2025.

\bibitem[Tian et~al.(2024)Tian, Gu, Li, Liu, Wang, Zhao, Zhan, Jia, Lang, and Zhao]{tian2024drivevlm}
Xiaoyu Tian, Junru Gu, Bailin Li, Yicheng Liu, Yang Wang, Zhiyong Zhao, Kun Zhan, Peng Jia, Xianpeng Lang, and Hang Zhao.
\newblock {DriveVLM: The Convergence of Autonomous Driving and Large Vision-Language Models}.
\newblock In \emph{Conference on Robot Learning (CoRL)}, 2024.

\bibitem[Wang et~al.(2025{\natexlab{a}})Wang, Xiong, Wang, and Chen]{wang2025bitvla}
Hongyu Wang, Chuyan Xiong, Ruiping Wang, and Xilin Chen.
\newblock {BitVLA: 1-Bit Vision-Language-Action Models for Robotics Manipulation}.
\newblock \emph{arXiv preprint arXiv:2506.07530}, 2025{\natexlab{a}}.

\bibitem[Wang et~al.(2025{\natexlab{b}})Wang, Luo, Bai, Cao, Che, Chen, Chen, Diamond, Ding, Ding, Feng, Heinrich, Huang, Karkus, Li, Li, Lin, Liu, Liu, Liu, Liu, Lu, Mao, Molchanov, Pavao, Peng, Ranzinger, Schmerling, Shen, Shi, Tariq, Tian, Wekel, Weng, Xiao, Yang, Yang, You, Zeng, Zhang, Ivanovic, and Pavone]{wang2025alpamayo}
Yan Wang, Wenjie Luo, Junjie Bai, Yulong Cao, Tong Che, Ke~Chen, Yuxiao Chen, Jenna Diamond, Yifan Ding, Wenhao Ding, Liang Feng, Greg Heinrich, Jack Huang, Peter Karkus, Boyi Li, Pinyi Li, Tsung-Yi Lin, Dongran Liu, Ming-Yu Liu, Langechuan Liu, Zhijian Liu, Jason Lu, Yunxiang Mao, Pavlo Molchanov, Lindsey Pavao, Zhenghao Peng, Mike Ranzinger, Ed~Schmerling, Shida Shen, Yunfei Shi, Sarah Tariq, Ran Tian, Tilman Wekel, Xinshuo Weng, Tianjun Xiao, Eric Yang, Xiaodong Yang, Yurong You, Xiaohui Zeng, Wenyuan Zhang, Boris Ivanovic, and Marco Pavone.
\newblock {Alpamayo-R1: Bridging Reasoning and Action Prediction for Generalizable Autonomous Driving in the Long Tail}.
\newblock \emph{arXiv preprint arXiv:2511.00088}, 2025{\natexlab{b}}.

\bibitem[Wen et~al.(2025)Wen, Zhu, Li, Zhu, Wu, Xu, Liu, Cheng, Shen, Peng, Feng, and Tang]{wen2025tinyvla}
Junjie Wen, Yichen Zhu, Jinming Li, Minjie Zhu, Kun Wu, Zhiyuan Xu, Ning Liu, Ran Cheng, Chaomin Shen, Yaxin Peng, Feifei Feng, and Jian Tang.
\newblock {TinyVLA: Towards Fast, Data-Efficient Vision-Language-Action Models for Robotic Manipulation}.
\newblock \emph{IEEE Robotics and Automation Letters}, 2025.

\bibitem[Xiao et~al.(2024)Xiao, Tang, Zuo, Guo, Yang, Tang, Fu, and Han]{xiao2024streamingllm}
Guangxuan Xiao, Yuandong Tang, Juntao Zuo, Junxian Guo, Shang Yang, Haotian Tang, Jinlong Fu, and Song Han.
\newblock {Efficient Streaming Language Models with Attention Sinks}.
\newblock In \emph{International Conference on Learning Representations (ICLR)}, 2024.

\bibitem[Xu et~al.(2025{\natexlab{a}})Xu, Wang, Xia, Zhu, Huang, and Xu]{xu2025vlacache}
Siyu Xu, Yunke Wang, Chenghao Xia, Dihao Zhu, Tao Huang, and Chang Xu.
\newblock {VLA-Cache: Efficient Vision-Language-Action Manipulation via Adaptive Token Caching}.
\newblock In \emph{Advances in Neural Information Processing Systems (NeurIPS)}, 2025{\natexlab{a}}.

\bibitem[Xu et~al.(2025{\natexlab{b}})Xu, Zhuang, and Shan]{xu2025kv}
Wanshun Xu, Long Zhuang, and Lianlei Shan.
\newblock {KV-Efficient VLA: A Method to Speed Up Vision Language Models with RNN-Gated Chunked KV Cache}.
\newblock \emph{arXiv preprint arXiv:2509.21354}, 2025{\natexlab{b}}.

\bibitem[Xu et~al.(2024)Xu, Zhang, Xie, Zhao, Guo, Wong, Li, and Zhao]{xu2024drivegpt4}
Zhenhua Xu, Yujia Zhang, Enze Xie, Zhen Zhao, Yong Guo, Kwan-Yee~K. Wong, Zhenguo Li, and Hengshuang Zhao.
\newblock {DriveGPT4: Interpretable End-to-End Autonomous Driving via Large Language Model}.
\newblock \emph{IEEE Robotics and Automation Letters}, 2024.

\bibitem[Yue et~al.(2024)Yue, Wang, Kang, Han, Wang, Song, Feng, and Huang]{yue2024deervla}
Yang Yue, Yulin Wang, Bingyi Kang, Yizeng Han, Shenzhi Wang, Shiji Song, Jiashi Feng, and Gao Huang.
\newblock {DeeR-VLA: Dynamic Inference of Multimodal Large Language Models for Efficient Robot Execution}.
\newblock In \emph{Advances in Neural Information Processing Systems (NeurIPS)}, 2024.

\bibitem[Zhang et~al.(2024)Zhang, Guo, Chen, Wang, Hu, Shi, and Chen]{zhang2024hirt}
Jianke Zhang, Yanjiang Guo, Xiaoyu Chen, Yen-Jen Wang, Yucheng Hu, Chengming Shi, and Jianyu Chen.
\newblock {HiRT: Enhancing Robotic Control with Hierarchical Robot Transformers}.
\newblock In \emph{Conference on Robot Learning (CoRL)}, 2024.

\bibitem[Zhang et~al.(2026)Zhang, Dong, Zhang, Heng, Chi, Dai, Du, Du, and Zhang]{zhang2026molevla}
Rongyu Zhang, Menghang Dong, Yuan Zhang, Liang Heng, Xiaowei Chi, Gaole Dai, Li~Du, Yuan Du, and Shanghang Zhang.
\newblock {MoLe-VLA: Dynamic Layer-Skipping Vision Language Action Model via Mixture-of-Layers for Efficient Robot Manipulation}.
\newblock In \emph{AAAI Conference on Artificial Intelligence (AAAI)}, 2026.

\bibitem[Zhou et~al.(2025)Zhou, Cai, Zhao, Zhang, Huang, Zhou, and Ma]{zhou2025autovla}
Zewei Zhou, Tianhui Cai, Seth~Z. Zhao, Yun Zhang, Zhiyu Huang, Bolei Zhou, and Jiaqi Ma.
\newblock {AutoVLA: A Vision-Language-Action Model for End-to-End Autonomous Driving with Adaptive Reasoning and Reinforcement Fine-Tuning}.
\newblock In \emph{Advances in Neural Information Processing Systems (NeurIPS)}, 2025.

\end{thebibliography}
\bibliographystyle{iclr2026}
}

\renewcommand{\thesection}{A.\arabic{section}}
\renewcommand{\thetable}{A\arabic{table}}
\renewcommand{\thefigure}{A\arabic{figure}}
\setcounter{section}{0}
\setcounter{table}{0}
\setcounter{figure}{0}

\newpage
\appendix
\section{Additional Experimental Results}

\subsection{Ablation}
\label{sec:ablation}

\paragraph{Streaming Fine-tuning}
\label{sec:finetune_vlm}
Fine-tuning the VLM does not recover the accuracy lost to streaming inference---it makes it worse (Tab.~\ref{tab:streaming_finetune}). As discussed in \S\ref{sec:streaming}, direct streaming introduces approximation errors into the KV cache, so we test whether fine-tuning the VLM can calibrate the cache and mitigate them. The results show the opposite: fine-tuning only the VLM substantially degrades performance, falling below even the setting without any fine-tuning. This further confirms that the streaming approximation mainly affects the action expert.

\paragraph{Speculative reasoning.} 
A larger draft block size does not improve latency. Since the average reasoning length across 100 randomly sampled clips is approximately 16 tokens, we also train draft models with a block size of 16. The larger block yields more accepted tokens per block ($\sim$8 tokens), but the increased draft and verification costs offset the gain, resulting in latency similar to a block size of 8 (Tab.~\ref{tab:spec_reasoning}).

\begin{table}[htb]
    \centering
    \renewcommand{\arraystretch}{0.9}
    \begin{subtable}[t]{0.36\linewidth}
        \centering
        \begin{tabular}{lcc}
        \toprule
                    & minADE\textsubscript{1}  & minADE\textsubscript{6} \\ \midrule
        Baseline    & 1.72 & 0.77   \\
        w/o f.t.    & 2.04 & 0.96   \\
        w/ f.t. VLM & 4.69 & 2.98   \\
        w/ f.t. AE  & \textbf{1.73} & \textbf{0.79}   \\ \bottomrule
        \end{tabular}
    \caption{Streaming fine-tuning strategies.}
    \label{tab:streaming_finetune}
    \end{subtable}
    \hfill
    \begin{subtable}[t]{0.6\linewidth}
        \centering
        \begin{tabular}{lcccc}
        \toprule
         & \multicolumn{2}{c}{RTX 4090} & \multicolumn{2}{c}{RTX PRO 6000} \\
         & B=8 & B=16 & B=8 & B=16 \\
        \midrule
        Latency                  & \textbf{208.6} & 215.2 & 159.4 & \textbf{154.4} \\
        Throughput               & \textbf{246.2} & 230.8 & 317.5 & \textbf{305.0} \\
        minADE\textsubscript{1}  & \textbf{1.56}  & 1.56  & 1.57  & \textbf{1.56}  \\
        \bottomrule
        \end{tabular}
        \caption{Draft-model block size (B).}
        \label{tab:spec_reasoning}
    \end{subtable}
    \caption{Ablation experiments. (a) Fine-tuning only the action expert (AE) nearly recovers baseline accuracy after streaming, whereas fine-tuning the VLM makes accuracy far worse than no fine-tuning at all. (b) Draft models with block size (B) 8 and 16 achieve comparable latency and accuracy.}
    \label{tab:ablation}
\end{table}

\subsection{FlashDrive on Alpamayo 1}
\label{sec:apmy_1}
We further validate FlashDrive on Alpamayo 1 (R1). As shown in Tab.~\ref{tab:speedup_breakdown_apmy1}, FlashDrive achieves a $4.5\times$ end-to-end speedup over the baseline while improving $\text{minADE}_{1}$ by 0.21\,m, at a 0.11\,m $\text{minADE}_{6}$ cost. The per-stage breakdown confirms that the algorithm-system co-design remains effective at every stage of the pipeline, reinforcing FlashDrive's role as a general-purpose inference framework for reasoning VLA models across generations of the same architectural family.

\begin{table}[htb]
\centering
\resizebox{\textwidth}{!}{%
\begin{tabular}{llllllll}
\toprule
 & \multicolumn{5}{c}{Latency (with 1 Trajectory) (ms)} & \multicolumn{2}{c}{Trajectory Error (m)} \\
\cmidrule(lr){2-6}\cmidrule(lr){7-8}
 & Encode & Prefill & Decode & Action & Total & minADE\textsubscript{1} $\downarrow$ & minADE\textsubscript{6} $\downarrow$ \\
\midrule
Alpamayo 1 & 86.4 & 162.3 & 262.5 \trpt{56.5} & 192.6 & 703.8 & 1.869 & 0.800 \\
+ FlashDrive & 12.0 & 46.9 & 49.6 \trpt{299.2} & 46.9 & 155.4 & 1.662 & 0.910 \\
\bottomrule
\end{tabular}%
}
\caption{FlashDrive achieves a 4.5$\times$ overall speedup with one trajectory on Alpamayo 1, improving minADE\textsubscript{1} at a 0.11~m minADE\textsubscript{6} cost---the gains transfer across model generations. The \trpt{throughput} next to the decoding time denotes token throughput (tokens/s) for that setting.}
\label{tab:speedup_breakdown_apmy1}
\end{table}

\subsection{AlpaSim Metrics}
\label{sec:alpasim_metrics}

Tab.~\ref{tab:alpasim_metrics} summarizes the closed-loop metrics used in Tab.~\ref{tab:alpasim}. Unless otherwise noted, event-based metrics are aggregated over time by taking the maximum, so a trajectory is counted as positive if the event occurs at any point during the rollout.

\begin{table}[htb]
\centering
\small
\begin{tabularx}{\textwidth}{p{0.18\textwidth}Xp{0.18\textwidth}}
\toprule
\textbf{Metric} & \textbf{Definition} & \textbf{Aggregation} \\
\midrule
Collision & Whether any front, rear, or lateral collision occurs between the ego vehicle and another actor's bounding box. & Max over time \\
Off Road & Whether the ego vehicle leaves the drivable road region, based on coverage of the ego polygon by the current or nearby lane regions and road-edge checks. & Max over time \\
Wrong Lane & Whether the ego heading differs from the lane centerline direction by more than $2\pi/3$. This metric can be noisy near intersections or ambiguous lane geometry. & Max over time \\
$P_{\text{rel}}$ & Relative progress obtained by projecting the ego centroid onto the ground-truth trajectory up to the current time. & Min over time \\
$D_{\text{traj}}$ & Distance from the ego centroid to the full ground-truth trajectory. & Max over time \\
$D_{\text{loc}}$ & Distance from the ego centroid to the time-aligned ground-truth location. & Max over time \\
Plan Dev. & L2 deviation between consecutive plans on their overlapping future timestamps, with exponentially larger weight on near-term waypoints. & Mean over time \\
Latency & Per-step rollout latency, including model inference, trajectory optimization, and simulator rendering. & Mean per step \\
\bottomrule
\end{tabularx}
\caption{Definitions of the AlpaSim closed-loop metrics reported in Tab.~\ref{tab:alpasim}.}
\label{tab:alpasim_metrics}
\end{table}

\subsection{Visualization}

\paragraph{Qualitative Visualization}
\begin{figure*}[t]
    \centering
    \includegraphics[width=\textwidth]{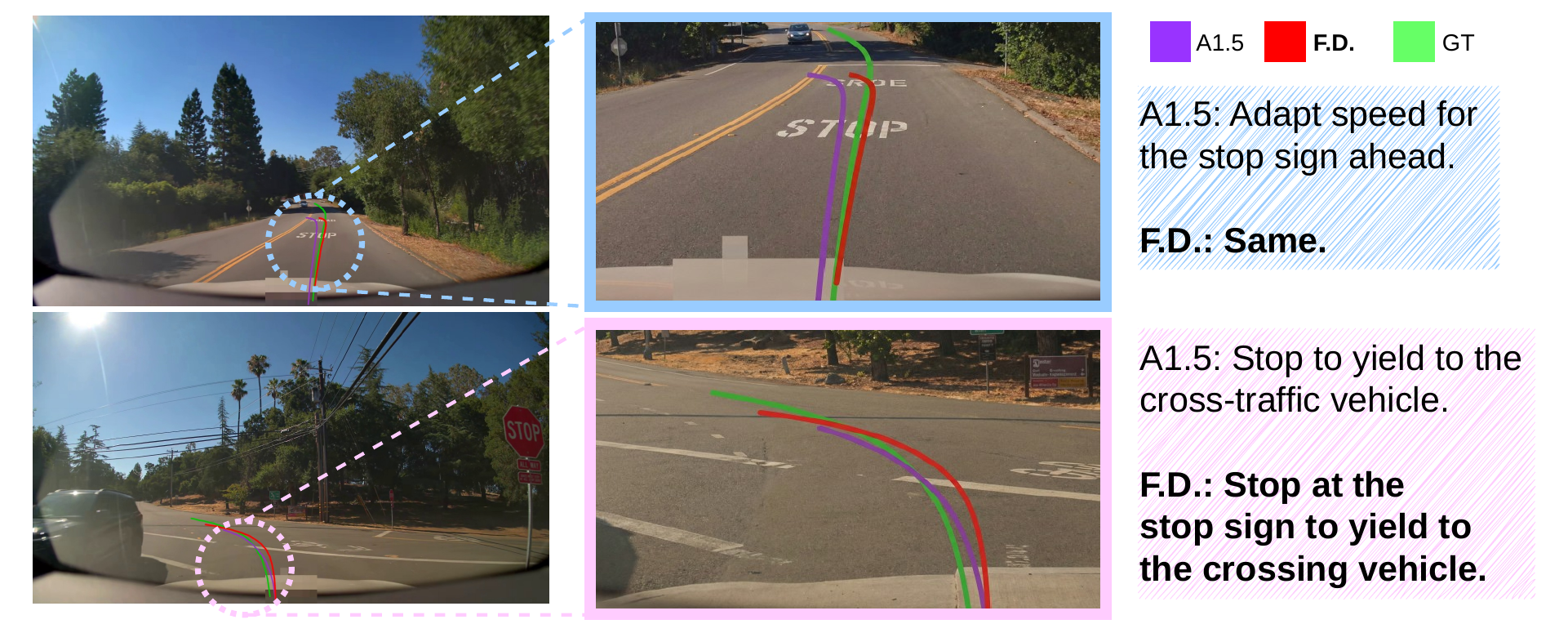}
    \caption{\textbf{Qualitative comparison of trajectories and chain-of-causation (CoC) reasoning.}
    FlashDrive's predicted trajectories align more closely with the ground truth than the baseline's, and its CoC reasoning remains consistent with the scene.
    \emph{(\textbf{Left})} shows the predicted trajectories,
    \emph{(\textbf{Right})} the corresponding CoC reasoning (Baseline vs. \textbf{FlashDrive (F.D.)}).
    Two common autonomous driving scenarios are selected for illustration:
    \textit{Longitudinal Straight Driving~({Row 1})}, and \textit{Oncoming Vehicle Encounter~({Row 2})}.}
    \label{fig:realworld_vis}
\end{figure*}

Fig.~\ref{fig:realworld_vis} presents a qualitative example of FlashDrive applied to the NVIDIA Autonomous Vehicle Dataset, compared against the Alpamayo 1.5 baseline. We overlay the predicted trajectories from Alpamayo 1.5 and FlashDrive alongside the ground truth. FlashDrive's trajectory aligns more closely with the ground truth, while its CoC reasoning tokens remain consistent with the current observation. We provide video visualizations on the project page (\url{https://z-lab.ai/projects/flashdrive})---we encourage readers to view them to appreciate FlashDrive's $4.7\times$ speedup in action.

\begin{figure}[htb]
    \centering
     \begin{subfigure}[b]{0.32\textwidth}
         \centering
         \includegraphics[width=\textwidth]{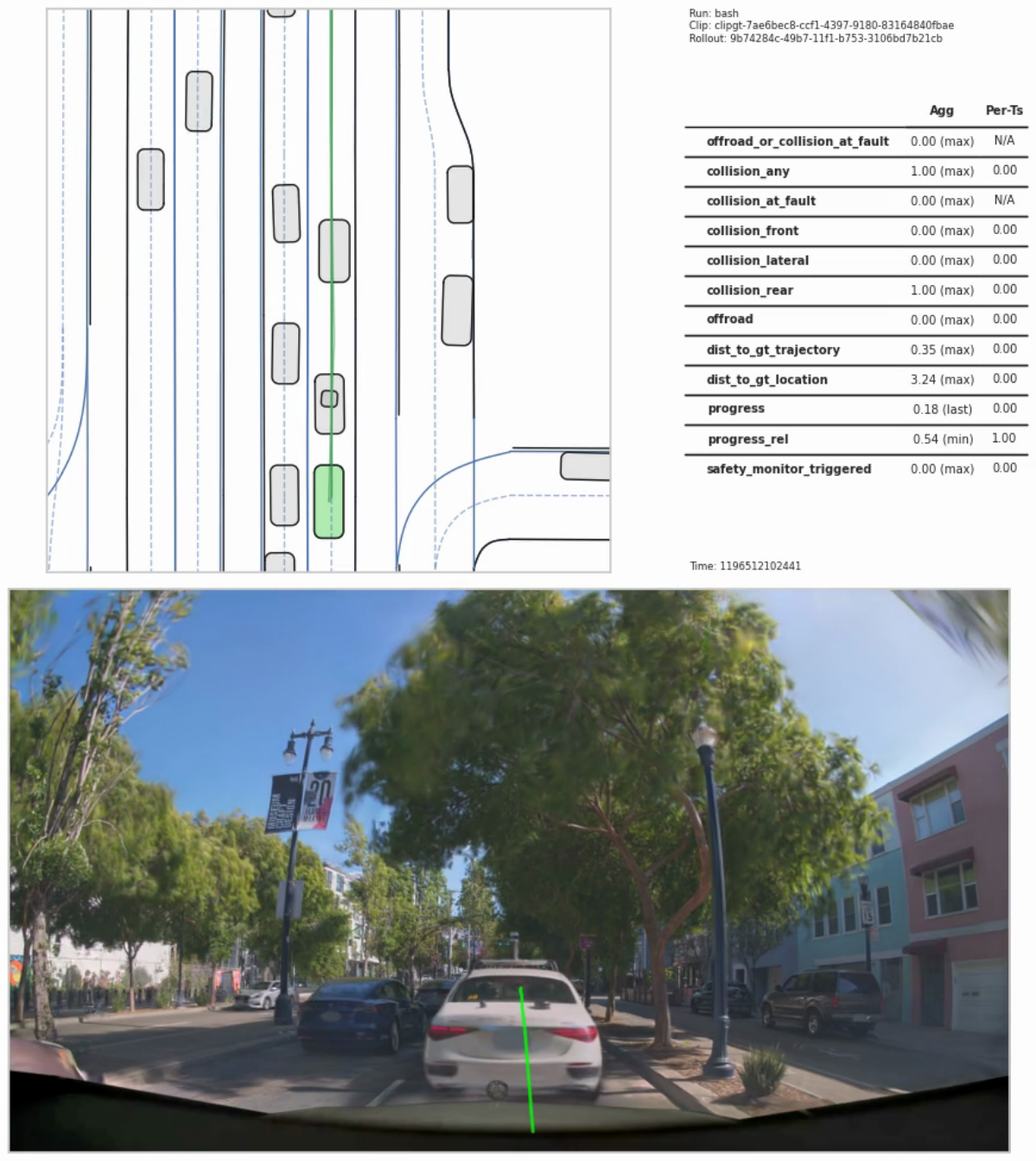}
     \end{subfigure}
     \hfill
     \begin{subfigure}[b]{0.32\textwidth}
         \centering
         \includegraphics[width=\textwidth]{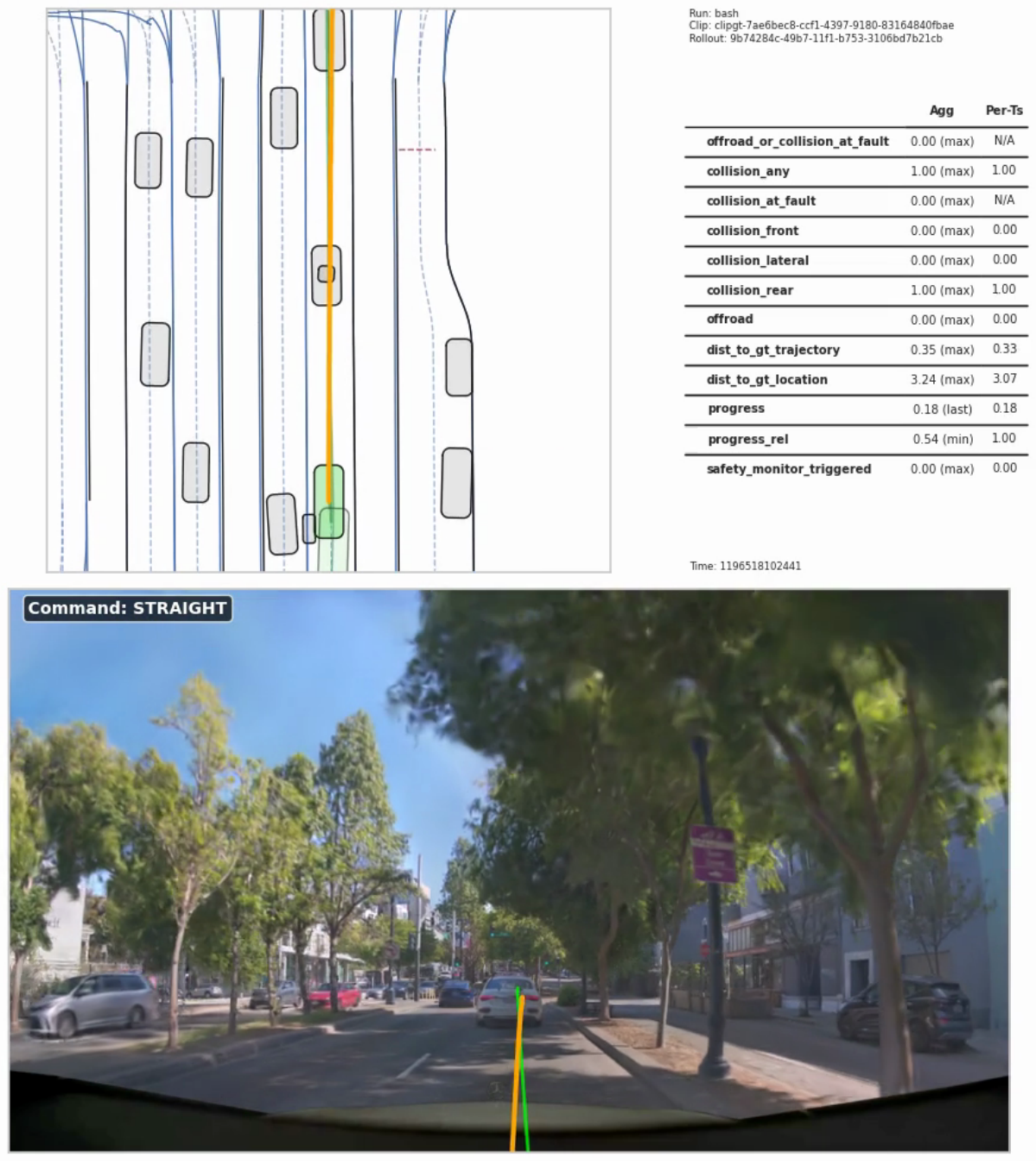}
     \end{subfigure}
     \hfill
     \begin{subfigure}[b]{0.33\textwidth}
         \centering
         \includegraphics[width=\textwidth]{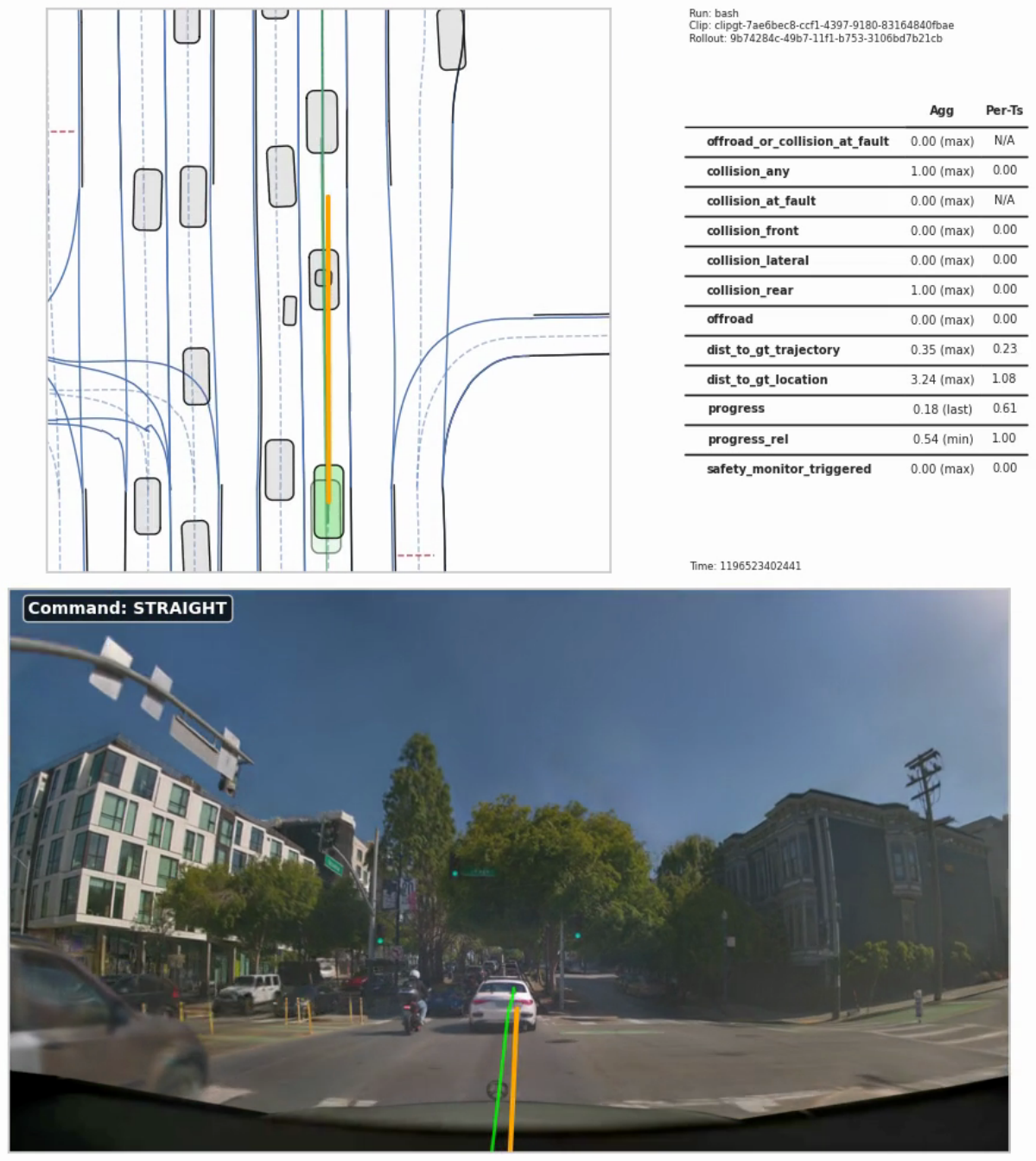}
     \end{subfigure}
     \caption{Closed-loop rollout of \method in AlpaSim. The accelerated policy tracks the route smoothly across the full episode.}
     \label{fig:alpasim_vis}
\end{figure}

\paragraph{AlpaSim closed-loop evaluation}
We present a rollout visualization from AlpaSim's closed-loop evaluation in Fig.~\ref{fig:alpasim_vis}. Video samples of FlashDrive's closed-loop rollouts are also available on the project page.

\end{document}